\documentclass[a4paper,10pt, twocolumn]{article}
\usepackage[utf8]{inputenc}

\usepackage{url}
\usepackage[numbered,framed]{mcode}
\usepackage{graphicx}
\usepackage{amsmath}
\usepackage{cleveref}
\usepackage{subfigure}
\usepackage{tikz}

\title{A tutorial on Particle Swarm Optimization Clustering}
\author{
    Augusto Luis Ballardini \\
    ballardini@disco.unimib.it}
\date{February 26, 2016}

\def\etal{\emph{et al. }}
\def\eg{\emph{e.g., }}
\def\ie{\emph{i.e., }}

\begin{document}

\maketitle

\begin{abstract}
This paper proposes a tutorial on the Data Clustering technique using the Particle Swarm Optimization approach. Following the work proposed by Merwe \etal\cite{1299577} here we present an in-deep analysis of the algorithm together with a Matlab implementation and a short tutorial that explains how to modify the proposed implementation and the effect of the parameters of the original algorithm.  Moreover, we provide a comparison against the results obtained using the well known K-Means approach. All the source code presented in this paper is publicly available under the GPL-v2 license.
\end{abstract}

\section{A gentle introduction}
What is Clustering? Clustering can be considered the most important unsupervised learning problem so, as every other problem of this kind, it deals with finding a structure in a collection of unlabeled data~\cite{Madhulatha2011}. We can also define the problem as the process of grouping together similar multi-dimensional data vectors into a number of clusters, or ``bins''~\cite{1299577}. According to the methodology used by the algorithm, we can distinguish four standard categories: Exclusive Clustering, Overlapping Clustering, Hierarchical Clustering and Probabilistic Clustering~\cite{mattClustering}. Along with these well-established techniques, several authors have tried to leverage the Particle Swarm Optimization (PSO)~\cite{488968} to cluster arbitrary data like, as an example, images. The contribution of Merwe and Engelbrecht~\cite{1299577} goes exactly along these lines, presenting two approaches for using PSO to cluster data along with an evaluation on six datasets and a comparison with the standard K-Means clustering algorithm.

While the reader can find an exhaustive description of the proposed algorithms on the original paper, in the rest of the work we will discuss our Matlab implementation of those algorithms together with a short but complete handbook for its usage. 

The remainder of this work is organized as follows. 
\Cref{sec:psopills} provides a brief introduction to the PSO technique and its formal definition.
\Cref{sec:psovskmeans} highlights the the benefits of PSO over state-of-the-art K-Means algorithm.
\Cref{sec:code} deals with the main key points of the Matlab code, providing the insights required to tailor the code to other datasets or clustering needs.

\begin{figure}[ht!]
  \centering
  \includegraphics[width=0.95\columnwidth]{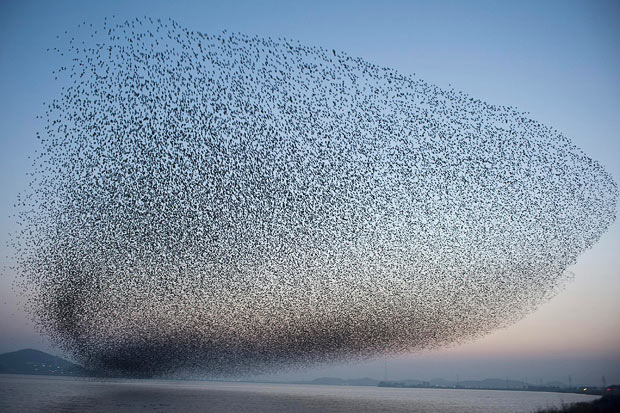}  
  \caption{The idea behind the PSO algorithm can be traced back to a group of birds randomly searching for food in an area. \label{fig1}}
\end{figure}

\section{The PSO algorithm in pills}\label{sec:psopills}
Particle Swarm Optimization (PSO) is a useful method for continuous nonlinear function optimization that simulates the so-called \emph{social behaviors}. The proposed methodology is tied to bird flocking, fish schooling and generally speaking swarming theory, and it is an extremely effective yet simple algorithm for optimizing a wide range of functions~\cite{488968}. The main insight of the algorithm is to maintain a set of potential solutions, \ie~\emph{particles}, where each one represents a solution to an optimization problem. Recalling the idea of bird flocks, a straightforward example that describes the intuition of the algorithm is described in~\cite{Xiaohui} and suppose a group of birds, randomly searching food in an area where there is only one piece of food. All the birds do not know where the food is but they know how far the food is in each time step. The PSO strategy is based on the idea that the best way to find the food is to follow the bird which is nearest to the food. 

\begin{figure}
  \centering
  \includegraphics[width=0.98\columnwidth]{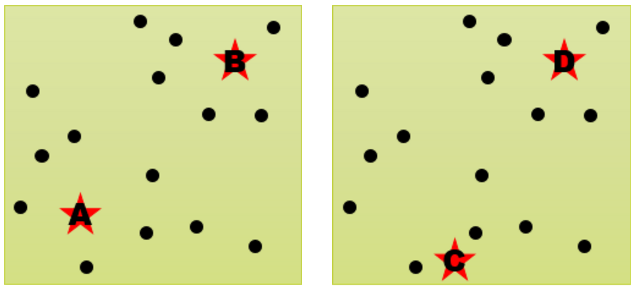}  
  \caption{In this example, we use two particles to cluster the given data using two clusters in 2 classes (dimensions). Each particle is represented using a green square, in which we can detect the two \emph{tracked} centroids (A and B in the first particle, C and D in the second particle). Please notice that the black dots represent the data, which is the same in each green square.\label{fig2}}
\end{figure}

\begin{figure}
  \centering
  \includegraphics[width=0.98\columnwidth]{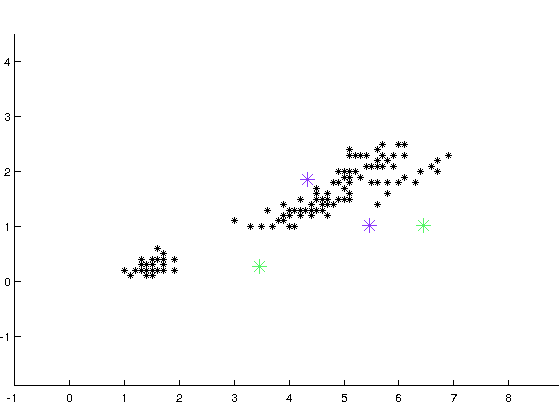}  
  \caption{The IRIS dataset initialized with two particles (green and magenta in this picture) each one using two $c$-centroids.\label{fig3}}
\end{figure}

Moving back to the context of clustering, we can define a solution as a set of $n$-coordinates, where each one corresponds to the $c$-dimensional position of a cluster centroid. In the problem of PSO-Clustering it follows that we can have more than one possible solution, in which every $n$ solution consists of $c$-dimensional cluster positions, \ie cluster centroids (see~\Cref{fig2,fig3}). It is important to notice that the algorithm itself can be used in any dimensional space, even though in the this work only 2D and 3D spaces are taken into account for the sake of  visualizing purposes. The aim of the proposed algorithm is then to find the best evaluation of a given fitness function or, in our case, the best spatial configuration of centroids. Since each particle represents a position in the $N_d$ space, the aim is then to adjust its position according to

\begin{itemize}
 \item the particle's best position found so far, and
 \item the best position in the neighborhood of that particle.
\end{itemize}

To fulfill the previous statements, each particle stores these values:

\begin{itemize}
 \item $x_i$, its current position 
 \item $v_i$, its current velocity
 \item $y_i$, its best position, found so far.
\end{itemize}

Using the above notation, intentionally kept as in~\cite{1299577}, a particle's position is adjusted according to:

\begin{equation}\label{eq:1}
\begin{split}
v_{i,k}(t+1) = \;\; & wv_{i,k}(t) + c_1r_{1,k}(t)(y_{i,k}(t)-x_{i,k}(t)) \\
& + c_2r_{2,k}(t)(y(t)-x_{i,k}(t))
\end{split}
\end{equation}

\begin{equation}\label{eq:2}
x_i(t+1) = x_i(t) + v_i(t+1)
\end{equation}

In~\Cref{eq:1} $w$ is called the \emph{inertia} weight, $c_1$ and $c_2$ are the acceleration constants, and both $r_{1,j}(t)$ and $r_{2,j}(t)$ are sampled from an uniform distribution $U(0,1)$. 
The velocity of the particle is then calculated using the contributions of (1) the previous velocity, (2) a \emph{cognitive} component related to its best-achieved distance, and (3) the \emph{social} component which takes into account the best achieved distance over all the particles in the swarm. The best position of a particle is calculated using the trivial~\Cref{eq:3}, which simply updates the best position if the fitness value in the current $i$-timestep is less than the previous fitness value of the particle.

\begin{equation}\label{eq:3}
y_i(t+1) = \left\{\begin{matrix}
y_i(t) & if \quad f(x_i(t+1)) \geq f(y_i(t)\\ 
x_i(t+1) & if \quad f(x_i(t+1)) < f(y_i(t)
\end{matrix}\right.
\end{equation}

The PSO is usually executed with a continuous iteration of the~\Cref{eq:1} and ~\Cref{eq:2}, until a specified number of iterations has been reached. An alternative solution is to stop when the velocities are close to zero, which means that the algorithm has reached a minimum in the optimization process.

One more time, it is important to notice that even if in~\cite{1299577} two kinds of PSO approaches are presented, respectively named \emph{gbest} and \emph{lbest} where the social components is basically bounded either to the current neighborhood of the particle rather than the entire swarm, in this work we refer only to the basic \emph{gbest} proposal. 

Before closing this section we need to introduce how to evaluate the PSO performance at each time step, \ie a descriptive measure of the fitness of the whole particle set.~\Cref{eq:4} implements this measure, where $|C_{i,j}|$ is the number of data vectors belonging to cluster $C_{ij}$, $z_p$ is the vector of the input data belonging the $C_{ij}$ cluster, $m_j$ is the $j$-th centroid of the $i$-th particle in cluster $C_{ij}$, $N_c$ is the number of clusters, and it can be described as follows. 

\begin{equation}\label{eq:4}
J_e = \frac{\sum_{j=1}^{N_c}\begin{bmatrix}
\sum_{ \forall Z\in C_{ij}}^{ }d(z_p, m_j)/|C_{i,j}|
\end{bmatrix}}{N_c}
\end{equation}

The ~\Cref{sec:code} will present an in-depth analysis of the code, where each step will be described within its relation with the formal definition just provided. 

\section{PSO vs K-Means}\label{sec:psovskmeans}

\begin{figure*}[htp!]  
  \centering
  \subfigure[Hybrid PSO approach]{\includegraphics[scale=0.48]{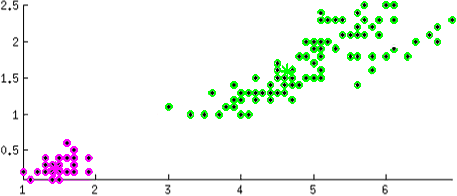}}\quad
  \subfigure[Standard PSO approach]{\includegraphics[scale=0.48]{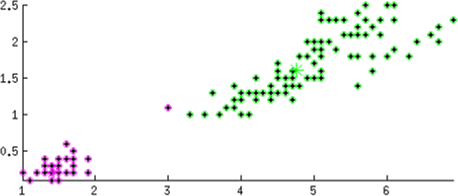}}
  \caption{The two figures depict the results of the PSO algorithm with and without the initial guess provided by K-Means (Hybrid-PSO). The reader can notice the slightly different results near the data point at coordinates (3,1), where the point is wrongly labeled as \emph{green} in the right figure while it is correctly assigned to the \emph{magenta} cluster in the left image. The data shown in this picture is retrieved from the IRIS dataset, considering only the first two features of the whole dataset.\label{fig4}}
\end{figure*}

Before moving on the code description, some important considerations should be highlighted. In particular, this section is focused to emphasize the benefits of PSO with respect to the well-known K-Means algorithm which, even if is one of the most popular clustering algorithms, shows as its main drawback its sensitivity to the initialization of the starting $K$ centroids. PSO tackles this problem by means of the incorporation of the three contributions stated in~\Cref{sec:psopills}, \ie inertia, cognitive and social components. It follows that the population-based search of the PSO algorithm reduces the effect that initial conditions have, since it starts searching from multiple positions in parallel and, even if PSO tends to converge slower (after fewer evaluations) than the standard K-Means approach, it usually yields to more accurate results~\cite{omran2002image}. As a final note, the performances of PSO can be further improved by seeding the initial swarm with the results of the K-Means algorithm, \eg using the results of K-Means as one of the particles and thus leaving the rest of the swarm randomly initialized. The latter approach, known as Hybrid-PSO and well-described in~\cite{1299577}, can effectively improve treacherous configurations like the one depicted in~\Cref{fig4}.

\section{The code, explained}\label{sec:code}

In this section, while mainly focusing on the key points of the PSO algorithm, \ie a detailed analysis of the meaningful code, will also highlight some tricky lines of the Matlab code, which may be not trivial for a novice Matlab user. In all the examples shown in this paper, we used the common Fisher's IRIS dataset~\cite{fisher1936use} provided in the Matlab environment, reducing its dimensionality to two or three in order to allow an easy visualization of the data and the clusters. Please notice that all the code lines provided in the listings correspond to the line numbers of the published Matlab code.

The code provides the following parameters:

\begin{lstlisting}[firstnumber=29, caption={The parameters of the proposed algorithm},label={lst:parameters}]
centroids      = 2 
dimensions     = 2
particles      = 2 
dataset_subset = 2
iterations     = 50
simtime        = 0.801
write_video    = false
hybrid_pso     = false
manual_init    = false
\end{lstlisting}

In the list, \emph{centroids} represent the number of clusters that the user wants to discover, \ie how many $n$-dimensional groups should be available in the input data, and corresponds to the $K$ value in the $K$-Means algorithm. The \emph{dimension} parameter specifies the $n$ value of each centroid that is, in a two-dimensional world, the $x$ and $y$ coordinates. Please note that these two values, \emph{centroids} and \emph{dimension}, are not mutually related as it is perfectly feasible to find two clusters in a three-dimensional space. The \emph{particles} parameter represents how many parallel swarms should be executed at the same time. Recall that each swarm, called also particle, represents a complete solution of the problem, \ie in the case of two centroids within a two-dimensional space, a couple two coordinates that localize the centroids. As an example, the user may refer to~\Cref{fig2}, where a set of two swarms is shown. The \emph{dataset\_subset} parameter allows to resize the original four-dimensional Matlab IRIS dataset to the specified value, allowing a \emph{2D} or \emph{3D} visualization. The meaning of the remaining parameters should be straightforward: \emph{iterations} simply counts how many times the algorithm will be reiterated before stopping its repetition, \emph{simtime} allows a pleasant visualization delay during the execution of the script, \emph{write\_video} enable the script to grab a video using as frames the image shown in each iteration, \emph{hybrid\_pso} seeds the PSO algorithm with the output of the standard Matlab K-Means implementation and the \emph{manual\_init} parameter allows, if the \emph{dimensions} parameter is set to $2$, to specify the initial position of the clusters. After this initial environment setup, the code provides three variables to specify the $w, c_1, c_2$ parameters of the~\Cref{eq:1} that control the inertial, cognitive and social contributions. In the code, these values were set according to~\cite{1299577, van2006analysis} to ensure a good convergence.

\begin{lstlisting}[firstnumber=41,caption={The specific PSO algorithm parameters},label={lst:PSO}]
w  = 0.72; %INERTIA
c1 = 1.49; %COGNITIVE
c2 = 1.49; %SOCIAL
\end{lstlisting}

\subsection{Assigning measures to cluster}

\begin{lstlisting}[firstnumber=176,caption={A non-trivial assignment},label={lst:meas2cluster}]
for particle=1:particles
  [value, index] = min(distances(:,:,particle),[],2);
  c(:,particle) = index;
end
\end{lstlisting}

\begin{figure*}[]  
  \centering  
    \subfigure[\label{fig5}]{\includegraphics[width=0.85\columnwidth]{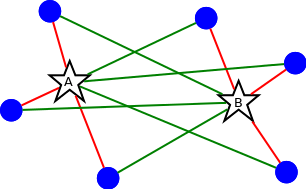}}\\
    \subfigure[\label{fig6}]{\includegraphics[width=0.85\columnwidth]{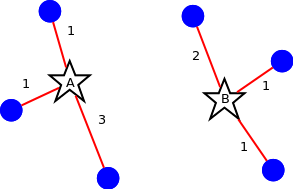}}\\
    \subfigure[\label{fig7}]{\includegraphics[width=0.85\columnwidth]{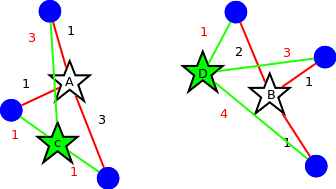}}  
  \caption{In this example, a dataset of with 6 data points (blue points) is clustered using two centroids. In (a) the distances to the closest centroid is marked in red. In (b) only the distance to the closest centroid is shown along with a measure of distance. Two averages, \ie \emph{local fitnesses} are then calculated using~\cref{matlabeq8_10}. In the case with multiple particles like in (c), where two swarms depicted with different colored stars are present, the process is iterated over every~\emph{particle} and a final value of \emph{global fitness} is chosen, selecting it from the minimum \emph{local fitness} set. Please note that in addition to the fitness value also the \emph{local} and \emph{global} positions are stored, respectively in~\cref{matlabeq8_5,matlabeq8_7} of~\Cref{lst:listingANY}.}  
  \label{fig8}
\end{figure*}

Implementing the~\Cref{eq:4} for calculating the fitness of a particle is trivial but the Matlab implementation may seem hard to understand at a glance. The code needed to calculate the fitness starts with~\cref{matlabeq8_1} checking if inside the array $c$ there is at least one element belonging to the $centroid$-th centroid. The local fitness is then defined as the mean of all the distances between the points belonging to each centroid. Since multiple particles can be evaluated in parallel, an additional loop is introduced in~\cref{matlabeq8_2}, allowing the code to iterate through the multiple swarm fitness evaluation. Please note that, during the second loop, we store and update two additional values in~\cref{matlabeq8_4,matlabeq8_5}, \ie the \emph{local best} fitness and position found so far, while at~\cref{matlabeq8_6,matlabeq8_7} we extract the \emph{very best} fitness value and position of all the particle swarm currently used. An overview of this process is shown in~\Cref{fig5}.

\begin{lstlisting}[firstnumber=216,caption={In this listing the code that controls the \emph{fitness} evaluation is reported. Please note that the global fitness is evaluated after the evaluation of the whole local fitness's set.},label={lst:listingANY}]
for particle=1:particles(*@\label{matlabeq8_2}@*)
  for centroid = 1 : centroids
    if any(c(:,particle) == centroid)(*@\label{matlabeq8_1}@*)
      local_fitness = ...
      mean(distances(c(:,particle)==centroid,centroid,particle));(*@\label{matlabeq8_10}@*)
      average_fitness(particle,1)=average_fitness(particle,1)... 
				  + local_fitness;
    end
  end
  average_fitness(particle,1) = average_fitness(particle,1) / ...
				centroids;
  if (average_fitness(particle,1) < swarm_fitness(particle))(*@\label{matlabeq8_3}@*)
    swarm_fitness(particle) = average_fitness(particle,1);(*@\label{matlabeq8_4}@*)
    swarm_best(:,:,particle) = swarm_pos(:,:,particle); %LOCAL BEST (*@\label{matlabeq8_5}@*)
  end                                                   %FITNESS
end

[global_fitness, index] = min(swarm_fitness); %GLOBAL BEST FITNESS (*@\label{matlabeq8_6}@*)
swarm_overall_pose = swarm_pos(:,:,index);    %GLOBAL BEST POSITION(*@\label{matlabeq8_7}@*)
\end{lstlisting}

The last part of the code concerns about updating the \emph{inertia}, \emph{cognitive} and \emph{social} components that contribute to set the \emph{velocity} of the particles. Apart from the \emph{inertia} component scaled using only the $w$ parameter, the others use the previously calculated \emph{best local} and \emph{global} positions, respectively for the cognitive and social aspects.  All of the components, added together, creates the so-called \emph{swarm velocity}, that is used to update the overall swam position. In~\cref{positions2,positions3} the $r1$, $r2$, $c1$ and $c2$ variables corresponds to the parameters defined in ~\Cref{eq:1}.

\begin{lstlisting}[firstnumber=47,caption={The code shows how update the position of the whole particle swarm},label={lst:PSOupdate}]
for particle=1:particles        (*@\label{positions1}@*)
  inertia = w * swarm_vel(:,:,particle);
  cognitive = c1 * r1 * ...  (*@\label{positions2}@*)
	      (swarm_best(:,:,particle)-swarm_pos(:,:,particle));
  social = c2 * r2 * (swarm_overall_pose-swarm_pos(:,:,particle)); (*@\label{positions3}@*)
  vel = inertia+cognitive+social;
  % UPDATED PARTICLE ...
  swarm_pos(:,:,particle) = swarm_pos(:,:,particle) + vel ; % .. POSE
  swarm_vel(:,:,particle) = vel;                            % .. VEL
end
\end{lstlisting}

\subsection{Replacing the IRIS dataset}
The provided code is tailored for the Matlab IRIS dataset with a specific configuration, meaning that the visualization part mainly works only with two-dimensional and three-dimensional input. This is achieved by resizing the original four-classes $150\times4$ IRIS dataset either by $150\times2$ or $150\times3$. We trivially resized it for visualization purposes only, since only two or three classes can be effectively shown in a graph. Tests based on the dataset provided in~\cite{Lichman:2013} shown the feasibility of using high dimensional dataset, \ie more than 3 classes, using both the available approaches with basic changes in~\Cref{load_1,load_2,load_3}.

\begin{lstlisting}[firstnumber=56,caption={How to load the input dataset},label={lst:load}]
load fisheriris.mat (*@\label{load_1}@*)
meas = meas(:,1+dataset_subset:dimensions+dataset_subset); (*@\label{load_2}@*)
dataset_size = size (meas); (*@\label{load_3}@*)
\end{lstlisting}

\subsection{Video Grabbing}
We put some extra lines in the code to allow an easy video grabbing. This feature is ensured by means of the \emph{getframe} and \emph{writeVideo} Matlab functions and their usage is trivial as follows. In the listing~\cref{videoGRAB_1,videoGRAB_2,videoGRAB_3} open the filesystem using as an output filename \emph{PSO.avi}, which will be located in the same folder of the Matlab Code.~\Cref{videoGRAB_4,videoGRAB_5} grab and insert an image in the video, while~\cref{videoGRAB_6,videoGRAB_7,videoGRAB_8} close the previously opened file.

\begin{lstlisting}[firstnumber=47]
    writerObj = VideoWriter('PSO.avi'); 	(*@\label{videoGRAB_1}@*)
    writerObj.Quality=100; 			(*@\label{videoGRAB_2}@*)
    open(writerObj); 				(*@\label{videoGRAB_3}@*)
\end{lstlisting}
\begin{lstlisting}[firstnumber=243,aboveskip=-5pt,belowskip=-5pt]
    frame = getframe(fh);			(*@\label{videoGRAB_4}@*)
    writeVideo(writerObj,frame);		(*@\label{videoGRAB_5}@*)
\end{lstlisting}
\begin{lstlisting}[firstnumber=295]
    frame = getframe(fh);			(*@\label{videoGRAB_6}@*)
    writeVideo(writerObj,frame);		(*@\label{videoGRAB_7}@*)
    close(writerObj);				(*@\label{videoGRAB_8}@*)
\end{lstlisting}

\section{Conclusions}
In this paper, a systematic explanation of the PSO-Algorithm proposed in~\cite{1299577} was presented by means of the analysis of the code publicly available at~\cite{ballardinicode}. The code provides both the standard PSO and the Hybrid-PSO options, allowing the user to master every detail of the original work. Although the code was originally tailored to be executed using the Matlab IRIS dataset, it can be easily adapted in order to perform clustering of potentially any kind of dataset with minimal code changes. 

\section*{Acknowledgement}
The author would like to thank Dr. Axel Furlan for his support in writing the first prototype of the following Matlab Code.

\onecolumn
\section{Matlab Code}
\begin{lstlisting}
% Author:  Augusto Luis Ballardini
% Email:   augusto.ballardini@disco.unimib.it
% Website: http://www.ira.disco.unimib.it/people/ballardini-augusto-luis/

% This library is distributed in the hope that it will be useful,
% but WITHOUT ANY WARRANTY; without even the implied warranty of
% MERCHANTABILITY or FITNESS FOR A PARTICULAR PURPOSE.
% Permission is granted to copy, distribute and/or modify this document
% under the terms of the GNU Free Documentation License, Version 1.3
% or any later version published by the Free Software Foundation;
% with no Invariant Sections, no Front-Cover Texts, and no Back-Cover Texts
% A copy of the license is included in the section entitled "GNU
% Free Documentation License".

% The following code is inspired by the following paper:
% Van Der Merwe, D. W.; Engelbrecht, AP., "Data clustering using particle
% swarm optimization," Evolutionary Computation, 2003. CEC '03. The 2003
% Congress on , vol.1, no., pp.215,220 Vol.1, 8-12 Dec. 2003
% doi: 10.1109/CEC.2003.1299577
% URL: 
% http://ieeexplore.ieee.org/stamp/stamp.jsp?tp=&arnumber=1299577

clear;
close all;

rng('default') % For reproducibility

% INIT PARTICLE SWARM
centroids = 2;       % == clusters here (aka centroids)
dimensions = 2;      % how many dimensions in each centroid
particles = 1;       % how many particles in the swarm, how many solutions
iterations = 50;     % iterations of the optimization alg.
simtime=0.101;       % simulation delay btw each iteration
dataset_subset = 2;  % for the IRIS dataset, change this value from 0 to 2
write_video = false; % enable to grab the output picture and save a video
hybrid_pso = true;   % enable/disable hybrid_pso
manual_init = false; % enable/disable manual initialization 
                     %                (only for dimensions={2,3})
                     
% GLOBAL PARAMETERS (the paper reports this values 0.72;1.49;1.49)
w  = 0.72; %INERTIA
c1 = 1.49; %COGNITIVE
c2 = 1.49; %SOCIAL

% VIDEO GRAB STUFF...
if write_video
    writerObj = VideoWriter('PSO.avi');
    writerObj.Quality=100;
    open(writerObj);
end

% LOAD DEFAULT CLUSTER (IRIS DATASET); USE WITH CARE!
% Resize the dataset with current "dimensions" variable. the standard iris
% dataset in matlab is 150x4, in this tutorial we need 150x2 or 150x3 for
% visualization purposes
load fisheriris.mat
meas = meas(:,1+dataset_subset:dimensions+dataset_subset); 
dataset_size = size (meas);

% Execute k-means if enabled the hybrid_pso approach. If enabled, the
% startint position of the pso algorithm will be initialized using the
% output of the standard matlab implementation of k-means.
if hybrid_pso
    fprintf('Running Matlab K-Means Version\n');
    [idx,KMEANS_CENTROIDS] = kmeans(meas,           ...
                                    centroids,      ...
                                    'dist',         ...
                                    'sqEuclidean',  ...
                                    'display',      ...
                                    'iter',         ...
                                    'start',        ...
                                    'uniform',      ...
                                    'onlinephase',  ...
                                    'off');
    fprintf('\n');
end

% PLOT STUFF... HANDLERS AND COLORS.
% This lines pre-configures the variables that will be used to plot the
% data.
pc = []; txt = [];
cluster_colors_vector = rand(particles, 3);

% PLOT DATASET
% This block creates either a 2d or 3d plot according to the "dimensions"
% variable. Please note that for visualization purposes the only admissible
% values are 2 (two) or 3 (three).
fh=figure(1);
hold on;
if dimensions == 3
    plot3(meas(:,1),meas(:,2),meas(:,3),'k*');
    view(3);
elseif dimensions == 2
    plot(meas(:,1),meas(:,2),'k*');
end

% PLOT STUFF .. SETTING UP AXIS IN THE FIGURE
% Reconfiguring the axis in the figure. Without this line the axis max/min
% values may change during runtime.
axis equal;
axis(reshape([min(meas)-2; max(meas)+2],1,[]));
hold off;

% SETTING UP PSO DATA STRUCTURES
% Here the variables needed in the pso clustering are pre-initialized.
% Please note that swarm_vel, swarm_pos and swarm_best maintains the values
% for all the swarms (aka particles)
% 'c' =
% 'ranges' is used to scale the initial randomized values to something
% inside the range of the input data (just to not have useless values
% outside the valid range, i.e. the range of the data).
% 'swarm_fitness' is initially set as infinite. this is the "value" that
% will become smaller and smaller (i.e. minimizating the fitness function)
swarm_vel = rand(centroids,dimensions,particles)*0.1;
swarm_pos = rand(centroids,dimensions,particles);
swarm_best = zeros(centroids,dimensions);
c = zeros(dataset_size(1),particles);
ranges = max(meas)-min(meas); % used to scale the values
swarm_pos = swarm_pos .* repmat(ranges,centroids,1,particles) + ...
                         repmat(min(meas),centroids,1,particles);
swarm_fitness(1:particles)=Inf;

% KMEANS_INIT
% Here, if the hybrid pso approach was selected, we replace the first
% swarm/solution to the result of the k-means algorithm. please note that
% even with this initialization the pso will somehow try to improve this
% guess, since the velocities of the swarm are still randomly set, meaning
% that the system is unstable at the very beginning.
if hybrid_pso
    swarm_pos(:,:,1) = KMEANS_CENTROIDS;
end

% MANUAL INITIALIZATION (only for dimension 2 and 3)
% For dimension 2 (two) we can add an user-initialization of the algorithm.
% This will eventually replace the k-means initialization, since here we
% replace again the first swarm/solution <notice the swarm_pos(:,:,**1**)>
% In the case of dimensions==3, i put here a random value, you can change
% these meaningless numbers without any problem <[6 3 4; 5 3 1]>
if manual_init
    if dimensions == 3
        % MANUAL INIT ONLY FOR THE FIRST PARTICLE (with 'random' numbers!)
             swarm_pos(:,:,1) = [6 3 4; 5 3 1]; 
    elseif dimensions == 2
        % KEYBOARD INIT ONLY FOR THE FIRST PARTICLE
             swarm_pos(:,:,1) = ginput(2);
    end
end

% Here the real PSO-algorithm begins
for iteration=1:iterations

    % CALCULATE EUCLIDEAN DISTANCES TO ALL CENTROIDS
    % Here we evaluate the distance (default 2-norm) between each centroid
    % inside each particle against all the values inside the input data
    % vector (the 'meas' variable resized in the very beginning). Keep all
    % the distances in the 'distances' variable.
    distances=zeros(dataset_size(1),centroids,particles);
    for particle=1:particles
        for centroid=1:centroids
            distance=zeros(dataset_size(1),1);
            for data_vector=1:dataset_size(1)
                %meas(data_vector,:)
                distance(data_vector,1) = norm( ...
                    swarm_pos(centroid,:,particle)-meas(data_vector,:));
            end
            distances(:,centroid,particle) = distance;
        end
    end
    
    % ASSIGN MEASURES with CLUSTERS 
    % using the 'min' Matlab function to find the "Smallest elements in
    % array" we create an 150xn matrix where the first column represents
    % the distances of each input value to neares current centroids, and
    % the n-columns specifies to which cluster/centroid the distance
    % refers to.    
    for particle=1:particles
        [value, index] = min(distances(:,:,particle),[],2);
        c(:,particle) = index;
    end

    % PLOT STUFF... CLEAR HANDLERS
    % clean the figure before plotting again
    delete(pc); delete(txt);
    pc = []; txt = [];
    
    % PLOT STUFF...
    % plotting again this step
    hold on;
    for particle=1:particles
        for centroid=1:centroids
            if any(c(:,particle) == centroid)
                if dimensions == 3
                    pc = [pc plot3(swarm_pos(centroid,1,particle), ...
                        swarm_pos(centroid,2,particle), ...
                        swarm_pos(centroid,3,particle),'*','color', ...
                        cluster_colors_vector(particle,:))];
                elseif dimensions == 2
                    pc = [pc plot(swarm_pos(centroid,1,particle), ...
                        swarm_pos(centroid,2,particle),'*','color',...
                        cluster_colors_vector(particle,:))];
                end
            end
        end
    end
    set(pc,{'MarkerSize'},{12})
    set(gca,'LooseInset',get(gca,'TightInset'));
    hold off;
 
    % CALCULATE GLOBAL FITNESS and LOCAL FITNESS:=swarm_fitness
    % Here I evaluate the fitness of the algorithm, measured as the
    % quantization error using the equation 8 of the original paper. It
    % also calculates the global best and local best positions using
    % equation 5. Please refer to the tutorial for explanation of this
    % equation.
    average_fitness = zeros(particles,1);
    for particle=1:particles
        for centroid = 1 : centroids
            if any(c(:,particle) == centroid)
                local_fitness = ...
                mean(distances(c(:,particle)==centroid,centroid,particle));
                average_fitness(particle,1)=average_fitness(particle,1)... 
                                            + local_fitness;
            end
        end
        average_fitness(particle,1) = average_fitness(particle,1) / ...
                                      centroids;
        if (average_fitness(particle,1) < swarm_fitness(particle))
            swarm_fitness(particle) = average_fitness(particle,1);
            swarm_best(:,:,particle) = swarm_pos(:,:,particle); %LOCAL BEST 
        end                                                     %FITNESS
    end    
    [global_fitness, index] = min(swarm_fitness);     %GLOBAL BEST FITNESS
    swarm_overall_pose = swarm_pos(:,:,index);        %GLOBAL BEST POSITION
    
    % SOME INFO ON THE COMMAND WINDOW
    % Here I print some info the the Matlab Command Window
    fprintf('%3d. global fitness is %5.4f\n',iteration,global_fitness);                
    pause(simtime);
            
    % VIDEO GRAB STUFF...
    % If the GRABBING option was selected, put the frame inside the video.
    if write_video
        frame = getframe(fh);
        writeVideo(writerObj,frame);
    end
       
    % SAMPLE r1 AND r2 FROM UNIFORM DISTRIBUTION [0..1]
    % Equation 3 and 4 needs a random value, sampled from an uniform
    % distribution. Here we go!
    r1 = rand;
    r2 = rand;
    
    % UPDATE CLUSTER CENTROIDS
    % Update the cluster centroids using equation 3 and 4. Here the
    % cognitive and social contributions are calculated to update the
    % velocity and position of each swar.
    for particle=1:particles        
        inertia = w * swarm_vel(:,:,particle);
        cognitive = c1 * r1 * ... 
                    (swarm_best(:,:,particle)-swarm_pos(:,:,particle));
        social = c2 * r2 * (swarm_overall_pose-swarm_pos(:,:,particle));
        vel = inertia+cognitive+social;
                
        % UPDATED PARTICLE ...
        swarm_pos(:,:,particle) = swarm_pos(:,:,particle) + vel ; % .. POSE
        swarm_vel(:,:,particle) = vel;                            % .. VEL
    end
    
end % end of the PSO algorithm

% PLOT THE ASSOCIATIONS WITH RESPECT TO THE CLUSTER 
% At the very end, paint the original points using the same color for the
% elements within the same cluster.
hold on;
particle=index; %select the best particle (with best fitness) 
cluster_colors = ['m','g','y','b','r','c','g'];
for centroid=1:centroids
    if any(c(:,particle) == centroid)
        if dimensions == 3
             plot3(meas(c(:,particle)==centroid,1),meas(c(:,particle)== ...
                centroid,2),meas(c(:,particle)==centroid,3),'o','color',...
                cluster_colors(centroid));
        elseif dimensions == 2
                plot(meas(c(:,particle)==centroid,1), ...
                meas(c(:,particle)==centroid,2),'o','color', ...
                cluster_colors(centroid));
        end
    end
end
hold off;

% VIDEO GRAB STUFF...
% Close the video file, if opened.
if write_video
    frame = getframe(fh);
    writeVideo(writerObj,frame);
    close(writerObj);
end

% SAY GOODBYE
fprintf('\nEnd, global fitness is %5.4f\n',global_fitness);
\end{lstlisting}

\newpage

\nocite{*}

\bibliographystyle{unsrt}
\bibliography{riferimenti}

\end{document}